# A deep neural network for multi-species fish detection using multiple acoustic cameras

FERNANDEZ GARCIA Guglielmo [1,2], MARTIGNAC François [1,2], NEVOUX Marie [1,2], BEAULATON Laurent [2,3], CORPETTI Thomas [4]

1) ESE, Ecology and Ecosystems Health, Institut Agro, INRAE, 35042 Rennes, France

2) Management of Diadromous Fish in their Environment, OFB, INRAE, Institut Agro, Univ. Pau & Pays Adour/E2S UPPA, Rennes, France

3) DRAS, OFB, 35042 Rennes, France

4) LETG-Rennes, Place du Recteur Henri Le Moal, 35043 Rennes Cedex, France

Corresponding Author : FERNANDEZ GARCIA Guglielmo, guglielmofernandez@gmail.com

Abstract : Underwater acoustic cameras are high potential devices for many applications in ecology, notably for fisheries management and monitoring. However how to extract such data into high value information without a time-consuming entire dataset reading by an operator is still a challenge. Moreover the analysis of acoustic imaging, due to its low signal-to-noise ratio, is a perfect training ground for experimenting with new approaches, especially concerning Deep Learning techniques. We present hereby a novel approach that takes advantage of both CNN (Convolutional Neural Network) and classical CV (Computer Vision) techniques, able to detect a generic class "fish" in acoustic video streams. The pipeline pre-treats the acoustic images to extract 2 features, in order to localise the signals and improve the detection performances. To ensure the performances from an ecological point of view, we propose also a two-step validation, one to validate the results of the trainings and one to test the method on a real-world scenario. The YOLOv3-based model was trained with data of fish from multiple species recorded by the two common acoustic cameras, DIDSON and ARIS, including species of high ecological interest, as Atlantic salmon or European eels. The model we developed provides satisfying results detecting almost 80% of fish and minimizing the false positive rate, however the model is much less efficient for eel detections on ARIS videos. The first CNN pipeline for fish monitoring exploiting video data from two models of acoustic cameras satisfies most of the required features. Many challenges are still present, such as the automation of fish species identification through a multiclass model.

However the results point a new solution for dealing with complex data, such as sonar data, which can also be reapplied in other cases where the signal-to-noise ratio is a challenge.

Keywords : ARIS, artificial intelligence, convolutional neural network, DIDSON, fisheries management, sonar camera

Running headline: Fish detection on sonar data with CNN

# Table of Content

1. Introduction

Recent technological innovations largely increase the capacities for scientists to collect at high frequency *in situ* data. Consequently, ecology, like other branches of the biological and Earth sciences, has entered a world of big data, and is confronting the attendant opportunities and challenges (Farley et al., 2018; Guo, 2017). To convert high-resolution data into high-value data, the development of optimized methodological approaches to extract relevant information in a time- and cost-effective way is a key step. This main challenge is common to all new numerical devices, in all ecosystems (Valletta et al., 2017). In aquatic environments, multi-beam high frequency acoustic or sonar cameras belong to the newest generation of hydroacoustic devices used for fish monitoring and species identification thanks to their high frequency and large detection beam (Foote, 2009; Martignac et al., 2015). These sonars record continuously data at very high frequency (e.g. up to 10 frames per second), leading to data storage issues and substantial time dedicated to data analysis (Martignac et al., 2015), but they can overcome many unsolvable operational limits of optical video-counting methods: the presence of a solid structure restricting the fish passages and an imperative light source, consequences of a very limited detection range. Acoustic cameras are relevant alternatives but the nature of the images is remarkably different from common underwater optical imaging (Kocak et al., 2008; Lu et al., 2017; Y. Wang et al., 2019). Although acoustic camera is efficient and high potential device to diverse applications in ecology science, the lack of solution for automatic analysis of the data (Helminen & Linnansaari, 2021; Shahrestani et al., 2017), greatly limits its deployment.

Computer Vision (CV) techniques, ranging from classical image analysis and machine learning (as Support Vector Machine, SVM, or Random Forests, RF) to deep learning techniques, have opened the possibility to automatically identify and classify objects of interest, offering tools that have displayed high efficiency and repeatability (LeCun et al., 2015). Such methods can answer the challenge of extracting valuables informations from big data also in ecology. Indeed, in recent years there has been a remarkable growth of the using of such methods for image data analysis (Christin et al., 2019; Kwok, 2019): from camera-trap species detection (Beery et al., 2018; Norouzzadeh et al., 2018; Schneider et al., 2018; Weinstein, 2018) and photo-identification of individuals (Miele et al., 2020) to video tracking (Bonneau et al., 2020), these methods are paving the way to automation also in the wide range of conditions that are of interest in ecology, such as night/day cycle, partial exposition of the object, different meteorological conditions. Most of these methods are developed on optical data (Weinstein, 2018), while sonar imaging possess several different features (Lankowicz et al., 2020) that makes the direct application of these models impossible and the development of an analysis tool a more challenging task: for examples low signal-to-noise ratio, a large spectrum of noise frequencies and a low resolution of the objects of interest. These characteristics already make sonar imaging challenging for automatic analysis, but each camera model exhibit slightly different features, adding the feat of creating multi-camera models. It has to be stressed that answering to these challenges might also lead to new insights on how to solve similar complex problems in ecology (e.g. noisy data). First experiments to automate the analysis have been conducted in other water-related fields such as AUV guidance systems (Zacchini et al., 2020) or underwater litter object detection (Valdenegro-Toro, 2016). Concerning fisheries management a few attempts have emerged recently (Christensen et al., 2020; Lee et al., 2018; Tarling et al., 2021). However the latter are limited or to single camera studies, or to synthetic data or to fish counting (without explicit detection): as we will see, this work represents therefore a first multi-camera, multi-species study, based on a large dataset (with a high variability data recorded on a long period of time), that poses a first step towards the development of general methods for acoustic cameras.

Machine learning algorithms require the calculation of some measurable property or characteristic of the images (so called "features") handly designed in advance, making the choice of the "good features" complicated (Jianbo Shi & Tomasi, 1994; Nixon & Aguado, 2019). On the contrary, deep learning algorithms are capable of calculating such "features" without any human intervention. Indeed, in recent years, important gains in performances have appeared in the community of image processing with the use of deep neural networks, whose objective is to construct a neural network composed of a large number of layers composed of elementary functions (neurons that model a simple relation between input/output). The concatenation of layers enables to model potentially very complex relations between inputs (noisy videos in our case) and outputs (detection of fish). Though the idea of combining several layers of neurons is old (H. Wang & Raj, 2017), the progresses in the recent years come from the fact that we have now enough data and associated computational resources to train such complex networks. In addition, some theoretical progresses on the definition and optimization of such networks have opened a wide range of applications. The reader can find in (LeCun et al., 2015) a general introduction to deep learning.

As for the processing of spatial data, the state of the art network for assigning a label to an image is the well-known CNN (Convolutional Neural Network). With this family of networks, a series of convolution and pooling layers jointly acts as automatic feature extractors that represent in a very relevant way the information contained in raw data. These features are so informative that complex computer vision tasks (classification, object detection, etc) become simpler. In recent years many architectures, either adapted to assign a label to each pixel (Fully Convolutional Networks and variants (Long et al., 2015)), to deal with unstructured data (Qi et al., 2017) or to time series (Karim et al., 2018) for example have been proposed. More recently, the community is also focusing on the fusion of complex data, as for example in the context of satellite images (Audebert et al., 2019).

In the context of acoustic imaging, because of the particular noise, the complexity of such data and their limited spatio-temporal variability, the feature extractor part of deep networks is still limited to distinguish fishes from noise, leading thus to poor performances. As a solution, we propose here to pre-process the data in a specific pipeline, based on Computer Vision, to help the optimization of the network to converge to a better minimum. The preprocessed data are then combined with the original one to provide "multichannel-layers", highlighting the interesting signal. This strategy, that is in general not common in deep learning, has already proven its results on optical underwater imaging (Salman et al., 2020) and should constitute an original but simple solution to the acoustic imaging problems.

Apart from showing the potential of the developed model, we aim also to provide recommendations for the implementation of this kind of approaches. Particular attention will also be given to the possibility for a technology transfer to users, by testing the model on an ecologically relevant dataset from different types of acoustic cameras. Reaching the goal of automatic data analysis will make possible in the future to count fish passage in real time, which will be an important tool for biodiversity protection and management structures to better understand the interactions between aquatic ecosystems and the human sphere.

## 2. Materials and Methods

### 2.1 Deep Learning Approach

We propose a hybrid method that embeds hand-crafted features in addition with raw images, to reduce the signal-to-noise ratio of the sonar data. By doing so, we inform the algorithm on interesting areas and therefore help the optimization on the network. Indeed, as it will be shown in Section 3.1, using raw images only leads to poor results. In practice we compute common CV features used in various domains such as image processing (Nixon & Aguado, 2019), physics (Raissi et al., 2019; Zdeborová, 2020) or geography (Chen, 2015; Ma et al., 2019). As for the network, we choose an architecture devoted to "object detection", that localises a bounding box and assigns a label to it. In this context the R-CNN (Girshick et al., 2014), that combines several CNN with selective search, enables to isolate regions of interest. In this work, we used the You Only Look Once (YOLO v3) developed by Redmon *et al.* (Redmon et al., 2016; Redmon & Farhadi, 2018). While other common architectures use a sliding trained classifier to detect object (as Faster R-CNN (Ren et al., 2016)), YOLO simultaneously predicts the position, the size and the probability for the object to belong to one class, leading to fast performances without losing accuracy.

Among the objectives of this work, we also aim at testing the feasibility of using well known neural networks, keep the model simple, and help detection by adding information about the image through pretreatment. Reasons why we choose to employ a well-known CNN in our pipeline. Our model should not be understood only as an efficient model for the automation of acoustic camera analysis, but also as an important step before the application of more complex architectures on this type of signal, such as LSTM+CNNs (Long short-term memory) (Ning et al., 2017; Xia et al., 2020).

### 2.2 Hydroacoustic methods: acoustic cameras

An acoustic camera is a multi-beam high-frequency sonar designed to create high-resolution images. Unlike common hydroacoustic methods, fish's skin and fins are better perceived by acoustic camera's very high frequencies, making centimetric fish measurement possible from the images (Martignac et al., 2015). According to their intrinsic characteristics (Table S.1), and being the first acoustic cameras available for ecological study purposes, DIDSON (Dual-frequency IDentification SONar; Sound Metrics Corp.) and ARIS (Adaptive Resolution Imaging Sonar; Sound Metrics Corp) are the most commonly used cameras to monitor fish in the wild (Cook et al., 2019; Helminen & Linnansaari, 2021; Keeken et al., 2021; Lagarde et al., 2020; Lenihan et al., 2020; Shahrestani et al., 2017; Yang et al., 2010).

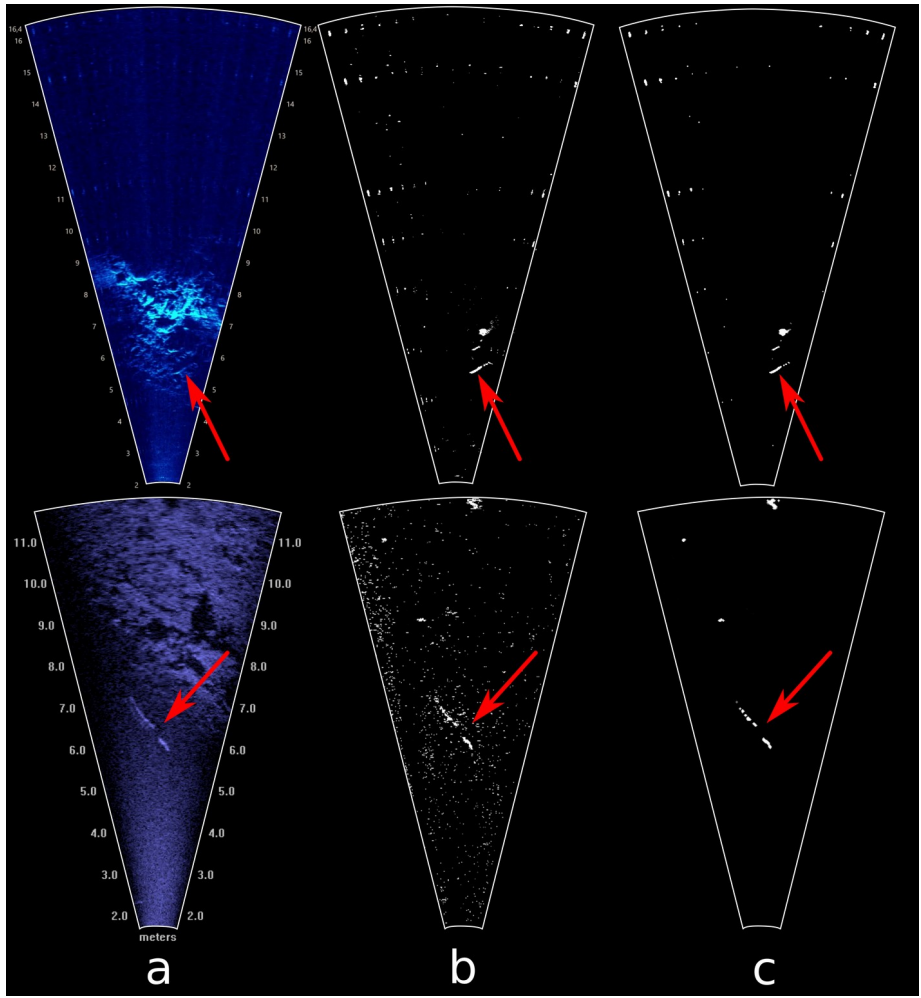

Figure 1. Examples of images (ARIS on top, DIDSON on bottom); raw images (a), background mask (b) and background filtered mask (c); the red arrow points to the fish position.

In the context of fish population monitoring in rivers, acoustic camera are horizontally oriented, set on one riverbank looking the other (Daum & Osborne, 1998; Maxwell & Gove, 2002). The acoustic datasets employed in this study have been recorded on monitoring sites during long-term and continuous surveys. Thus, the datasets include all the aquatic fauna activity in a multi-specific population with several species of interest in term of conservation and management of populations, such as diadromous or invasive fish species. DIDSON videos have been recorded on the Sélune River (Normandy, France) and ARIS videos on both Sélune and Touques rivers (Normandy, France). Examples of DIDSON and ARIS images are reported in Figure 1, panel a. The acoustic videos are sequences of contiguous images of more than 15 minutes. For both cameras, fish passages have been recorded at distances ranging from 1 to 17 meters, at a 1.8 MHz frequency (HF for DIDSON; LF for ARIS) at 5 frames per second. The fish species have been identified by an experienced operator on the video datasets, interpreting behavioural and morphological fish characteristics, adding biological expertise such as diadromous fish phenology or trophic behaviours (Martignac et al., 2015). These videos were used as a basis to elaborate both datasets.

2.3 Datasets

Two datasets have been used in this study: the "Training Dataset" (TD) to train, evaluate and test the deep learning model on a dataset of images and the "Validation Dataset" (VD), to test the full pipeline on videos with a high abundance of fish passages (more details on the datasets are given in SI).

The TD is composed of 4 817 images, 1 612 from ARIS camera's clips (short sequences where only the images with fishes have been chosen) and 3 205 from DIDSON camera' clips. The clips have been carefully chosen to integrate variability in the TD: different species, thus individual with large ranges of morphology (body shape, undulation, lengths) and behaviour (i.e. swimming pattern), different environmental conditions (i.e. river flow) and extracted from two monitoring sites, different recording window lengths and ranges, different times of year. We ensured that passages of fish species of interest were recorded in the TD: diadromous species, such as the Atlantic salmon (*Salmo salar*), the European eel (*Anguilla anguilla*), the sea lamprey (*Petromyzon marinus*), the Allis shad (*Alosa alosa*) and an opportunistic predator, the European catfish (*Silurus glanis*). In the TD, all the images have been annotated with a bounding box and a label corresponding to the species, even if all the labels coalesce during the treatment in a unique "fish" label (see Table S.2 for the distribution of each species in the TD). In order to ensure the accuracy of the model also on sequences where no passage are present, 1 196 images where no fish are present have been added to the test set (428 for the ARIS camera, 768 for the DIDSON camera).

The VD is composed of 184 videos (sequences of more than 15 minutes), 80 for ARIS camera and 104 for DIDSON camera, for a total of around 40 hours of recording. The videos of the VD integrate the same variability as the TD. Each fish larger than 20 cm have been counted and described with the passage hour (20 cm been, according our experience, the minimum size limit to efficiently see and track a fish), the fish size range (four classes: 20-40 cm, 40-60 cm, 60-80 cm and larger than 80 cm), its position in the detection beam, its swimming direction and its fish species if an undoubted identification was possible according its morphology and behaviour (otherwise a "generic fish" label was used), no bounding boxes were saved (see Table S.3 for the distribution of each species in the VD). In 33 clips (6 for ARIS camera and 27 for DIDSON camera), not a single fish passage is present, for a total of around 6 hours. More details on the datasets are given in S.1.

2.4 Deep Learning Model and Validation

During the training, the whole set of weights was optimised, starting from the pre-trained Darknet's weights of the original YOLO (Redmon et al., 2016). All training were performed with the same hyperparameters: learning rate of 0.001, decay of 0.0005 and momentum of 0.9. The network resolution was set to 608 x 608 pixels. For more details on the optimization of neural networks, we refer the reader to (R. Sun, 2019).

We defined two levels of validation: a "model validation", with a set of metrics commonly used to evaluate CNN models and calculated on the TD dataset, and an "ecological validation", in which the CNN model is tested on the VD dataset to assess the efficiency of the model in the frame of ecological studies and to conclude about the real-applicability on monitoring sites data.

$$\text{Confusion Matrix} = \begin{pmatrix} TP, FP \\ FN, TN \end{pmatrix}$$

Equation 1. Confusion matrix; TP is the number of true positives, FP of false positives, FN of False positives and TN of True Negative

Concerning the model validation, we adopted six metrics: precision, recall, $F_1$-score, AP@0.50, Cohen's kappa coefficient (explained in Table S.4) and confusion matrix. The latter is defined in Equation 1, where TP is the number of true positives images, FP false positives, FN False positives and TN true negative. The confusion matrix is normalised by column, so TP+FN=1 and FP+TN=1. We expect a confusion matrix that presents at least a TP rate greater than 70%-75% and a FP rate as low as possible, to avoid the operator spending too much time verifying the detections found. As mentioned in Section 2.2, a large part of the video stream consists of images with no fish (TN). Apart from the confusion matrix, to evaluate the equilibrium between TN and TP we have included in the metrics the Cohen's kappa coefficient, κ (Cohen, 1960; Landis & Koch, 1977). The latter already provides an interpretative framework, with a value considered "good" starting in general from 0.60-0.80.

Concerning the "ecological validation", the model is applied on the entire VD to assess its efficiency to detect fish larger than 20 cm. The entire reading of the VD videos' by an operator are compared to the model outputs to validate if the model detects efficiently the passage of one fish described by an operator. Even if the classifier is single-class, the results have been evaluated in terms of all species of interest. The comparison aims to assess the possible effects of biological and environmental variables on the model efficiency, such as the fish species (by their morphological and behavioural differences), the fish size range, distance to the camera, or the seasonal environmental conditions. When evaluating the VD, a filter has been added at the end of the pipeline, to avoid flash detections: all detections that do not have a subsequent detection in the next image with an overlapping bounding box are automatically rejected. Since this validation deals with sequences, we defined as a TP each passage of a fish for which the neural network have found the fish on at least two consecutive images, to avoid counting flash detections (detection on a single image). Indeed a fish passage can be identified by the detection on only a few consecutive images, so TP and FN can be easily defined in function of detected fishes. Such reasoning cannot be applied for the calculation of TN, expressed in number of images where no fish is found compared to the total number of images without any fish. This limitation of our approach makes impossible to write a confusion matrix where all elements are expressed in function of the same quantity. To evaluate the TN%, we took into account only the 33 clips without any individual present. Finally the FP are defined as each false detection of a fish consecutive on at least two images. We compared the FP number with the entire number of passages in the ground truth on the same clip. Such estimation gives an idea of how many detections an operator have to manually verify for each clip and consequently appear as a proxy of the operator time spending. In general, we aimed to maximize the TN% (to sort the empty sequences), but also to minimize the undetected fish passages, meaning a high TP% (larger than ~75%) and a rate between FP/TP as low as possible.

2.5 Data preprocessing

The first preprocessing step was the background segmentation of each clip, using the Gaussian Mixture-based Background/Foreground Segmentation Algorithm as implemented by Zivkovic *et al.* (Zivkovic & van der Heijden, 2006), to generate the "background masks", "b". The variance threshold for the pixel-model match depends on the camera type and were set to 130 for DIDSON clips and 10 for ARIS clips. Due to the noisy nature of the signal, the parameters were chosen to optimise noise reduction limiting meaningful loss of the signals. In order to further reduce the salt-and-pepper noise in the "b" masks, each one has been smoothed using a median filter with a 3x3 aperture (Huang et al., 1979). The resulting images were further processed with an opening operator (Dougherty & Lotufo, 2003), with a structuring element of dimensions 3x3 and a cross-hair shape. The resulting masks are named "background filtered masks", "$b_f$" (see Figure 1). The two masks and the raw images (called "r") are used to compose the processed RGB images "$rbb_f$". All the information "r" images are in the blue channel, therefore only this channel was used. The green channel was assigned to the "b" mask and the red channel to the "$b_f$" masks. To demonstrate the influence of each mask, we tested also the 2-channel images "rb" and "$rb_f$". During in the training, for all the models (with exception of the raw's one), colour data augmentation was disabled. The choice of these masks was driven by a "best performance" principle: several other sets of masks have been tested, ranging from temporal median filters to optical flow maps, with some of them converging to the same type of results (not presented). However a specific discussion of these sets of masks is beyond the scope of this paper.

## 3. Results

### 3.1 Influence of the data treatment

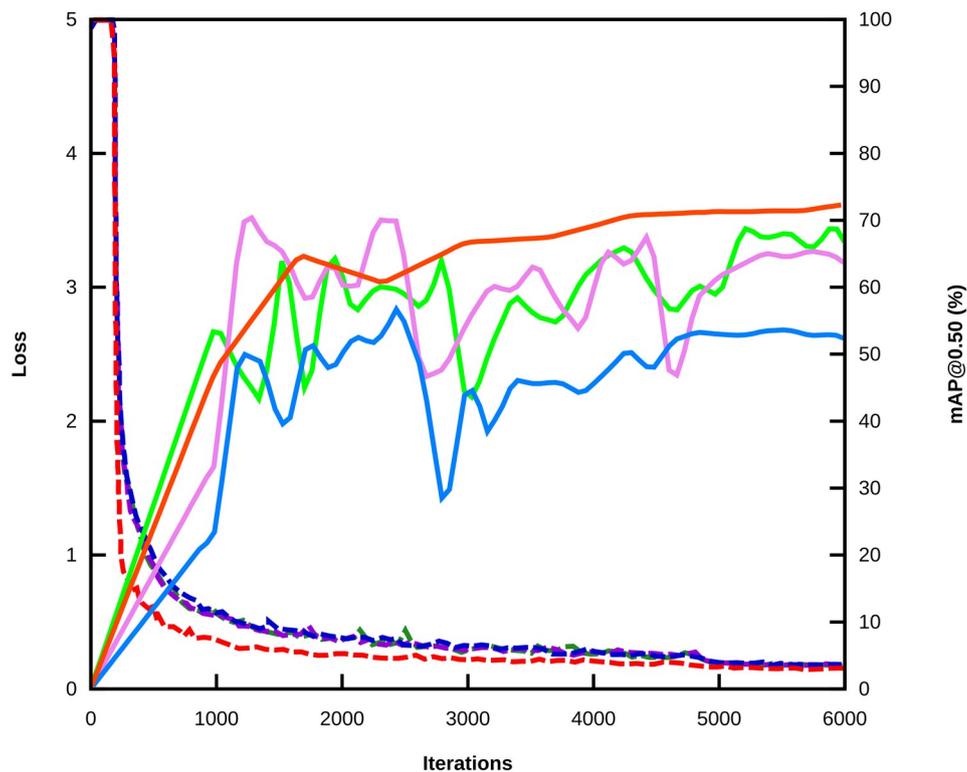

Figure 2. Loss (dotted line) and AP (solid line) for different models: r (blue), rb (green), rb$_f$ (violet) and rbb$_f$ (red).

As this is the first work datasets, multiple cameras and approach, it was necessary to assess the Deep Learning architecture on the different level of preprocessing. Amongst the different performance measures, we choose to compare the models using the AP@0.50 metric evaluated on the "val" set (see SI for details on the dataset splitting). As shown in Figure 2, the models trained on the unprocessed images and in combination of a single mask present an oscillating behaviour of the AP@0.50, suggesting an overfitting, and stabilise only at the end of the training. The model with unprocessed images ("r") reaches an AP@0.50 of 53%: this data constitutes, so, the "null model" for the others. Both the "rb" images and "rb$_f$" images exhibit a similar behaviour, showing an improvement in the evaluation metric that reaches 69% for the "rb" model. This data suggest that both the "b" mask or the "b$_f$", separately, improve the performance of the detection, but also that these masks individually are not a viable alternative to the original images (see Figure S.1). Finally the model trained on the "rbb$_f$" images shows stable convergence and is able to reach 72% of

AP@0.50. In order to verify which model is able to well generalise (being able to classify data of the same kind of images as the training ones, but that the CNN has never seen before), we compared the evaluation metrics (precision, recall, $F_1$-score and AP@0.50) for the different models on the "test" set of the TD dataset (see Table 1 and Figures S.1 and S.2). The AP@0.50 presents lower values for all the models, as expected, but surprisingly the "rb" model exhibits metrics similar to the "rbb$_f$", thus suggesting that in the latter the "b" feature is the most important one. The highest $F_1$-score is found for the "rbb$_f$" model: this, added to the well-behaved training showed in Figure 2, leaded to the choice of the "rbb$_f$" model (a scheme of this model is given in Figure 3).

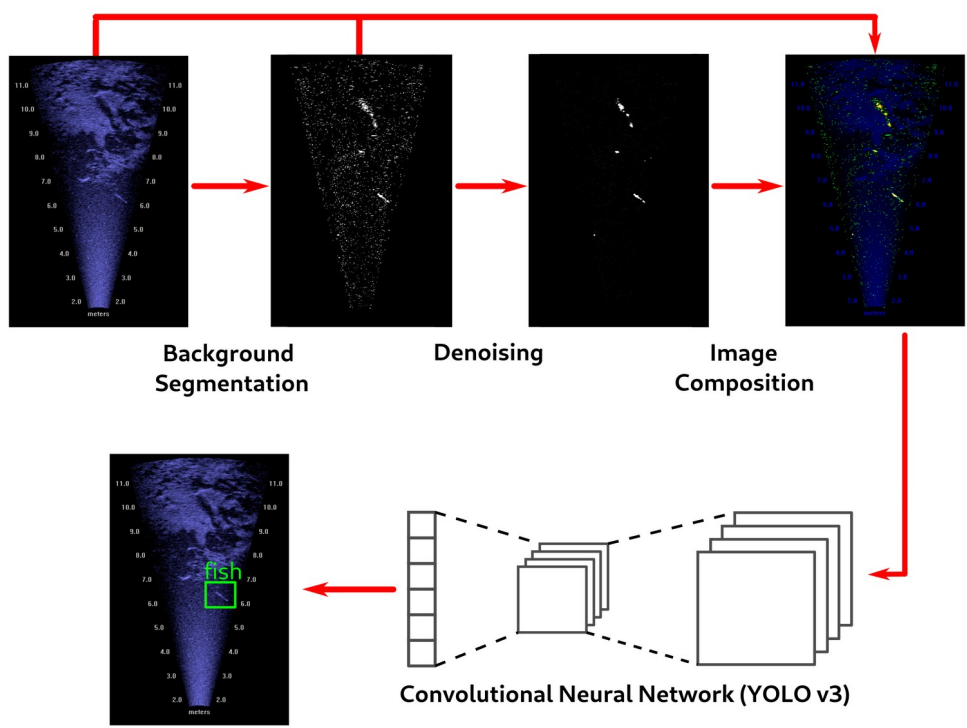

Figure 3. Scheme of the "rbb$_f$" model's pipeline, from raw data ("r") to the detections, through the background segmentation ("b" mask), the denoising step ("b$_f$" mask) and the CNN network.

|  | r | rb | rb$_f$ | rbb$_f$ |
|---|---|---|---|---|
| Precision | 0.71 | 0.71 | 0.78 | 0.78 |
| Recall | 0.41 | 0.65 | 0.50 | 0.61 |
| $F_1$-score | 0.52 | 0.67 | 0.61 | 0.69 |
| AP@0.50 | 37.64 | 62.43 | 53.66 | 64.57 |

Table 1. Validation metrics for the various models calculated on the "test" set of the TD dataset.

3.2 Model validation

As mentioned in Section 2.4, the focus of this work is not only on the detection of TP (of which some examples are showed in Figure 4), but also to reach good performances on TN/FN detections. The Cohen's kappa coefficient, κ, was calculated for the model trained on the "rbb$_f$" images on the "test" set of the TD dataset, leading to a κ = 0.73: for similar tasks (Fleiss et al., 1969) such score is largely above the threshold of the "good" score. This suggest that the detector is able to discriminate between TP and TN. To confirm this, we calculated the confusion matrix: as showed in Table 2, the detector is performing very well (87%) on the TN, while keeping good results (61%) for TP. Such result suggests that our detector is able to sort out efficiently all the sequences without any passage of fish, while maintaining a high probability to identify the sequences with the presence of an individual, even looking at the single image and not to a sequence. The confusion matrices of each camera (Table S.5) show different performances: while for DIDSON we reach optimal scores (larger than ~70%) for both TP% and TN%, for ARIS the TP% is too low, at least for such test based on single images.

|  |  | Ground truth | |
|---|---|---|---|
|  |  | Fish | No Fish |
| Predicted | Fish | 0.61 | 0.13 |
|  | No Fish | 0.39 | 0.87 |

Table 2. Confusion matrix for the "rbb$_f$" model, normalised by column, calculated on the "test" set of the TD dataset.

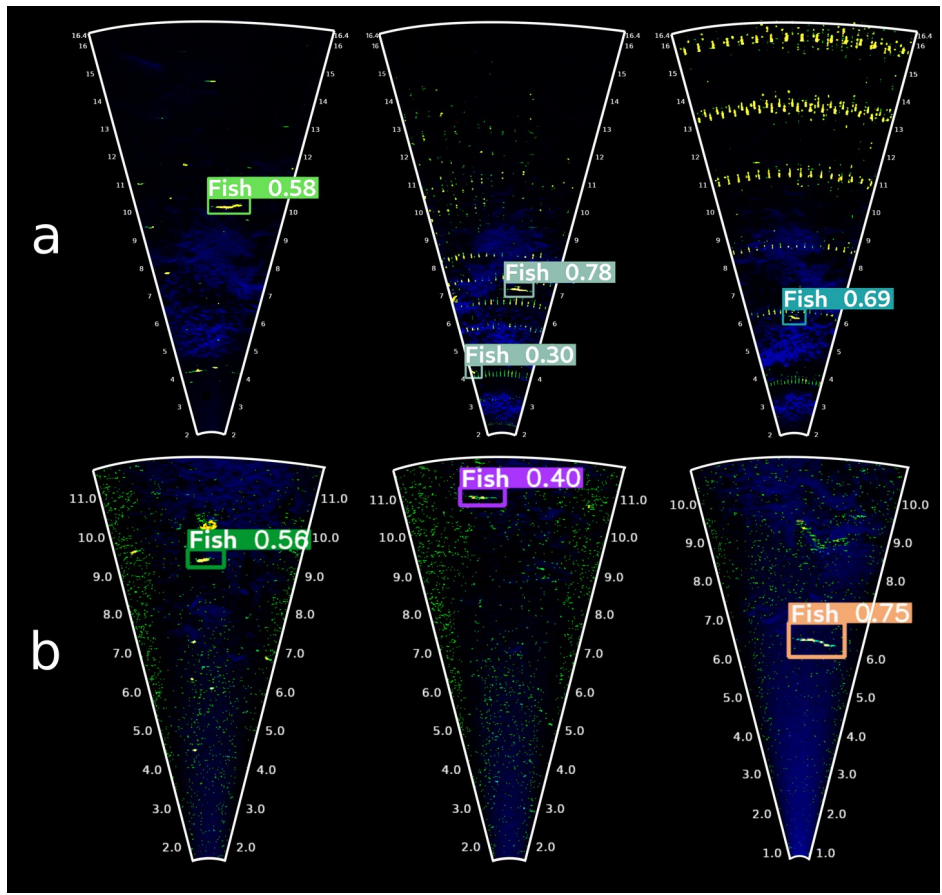

Figure 4: example of detections on the rbb$_f$ images from the test set of the TD dataset; ARIS images on top (a) with the detection of (from left to right) an Atlantic Salmon, a Sea Lamprey and a Giant catfish; DIDSON images on bottom (b) with the detection of (from left to right) a Sea Lamprey, Atlantic Salmon and a European catfish.

3.3 Ecological validation

As shown in Table 3, the total recall, or TP%, for the DIDSON camera exhibit an increasing trend with the fish size, with the best results starting from the 40-60 cm class and reaching more than 90% for large size fish. The same trend is present for the ARIS camera, but with lower results, with a maximum around 80% for large size fish. For small and medium-sized fish and for fish detected on the ARIS camera, the results are below the threshold of ~70-75%, for which the signal can be confused with the noise of sonar signal. Even if these results outline the general efficiency of our method, a detailed analysis of the main species of interest (Atlantic salmon and European eel) reveals some differences in the performances. For the Atlantic salmon, the results are remarkably high with both cameras and for all fish sizes, even if for the ARIS camera no large dataset is available. On the contrary, for the European eel the DIDSON camera offers satisfactory results (with TP larger than ~70%) only for 60-80 cm size. For the ARIS camera, none of the values exceed 75%. The FN% is simply 1-TP%, therefore the same observations apply: higher values of FN% are concentrated on small sizes, meaning that small individuals will not be efficiently detected by our automatic analysis.

|  |  | Recall (%TP) in function of fish size ||||  |
|---|---|---|---|---|---|---|
| Species | Camera | 20-40 | 40-60 | 60-80 | > 80 | Total |
| All species | DIDSON | 57.62 | 76.22 | 90.00 | 93.33 | 67.91 |
|  | ARIS | 52.87 | 55.39 | 83.72 | 88.89 | 55.13 |
| Atlantic salmon | DIDSON | NA | 92.86 | 96.97 | 100.00 | 96.23 |
|  | ARIS | NA | 100.00 | 100,00 | NA | 100.00 |
| European eel | DIDSON | NA | 57.14 | 93.33 | 66.67 | 75.00 |
|  | ARIS | 7.69 | 25.00 | 66.67 | 50.00 | 30.49 |
| Other species of interest | DIDSON | 0 | 100.00 | 100.00 | 83.33 | 94.44 |
|  | ARIS | NA | NA | NA | 71.43 | 71.43 |

Table 3. Recall (%TP) for the $rbb_f$ model, calculated on the VD dataset (dataset of videos).

For both cameras TN%, calculated on a set of empty long clips as explained in 2.4, is very high (around 99%), hinting that the method is capable to identify the sequences where an individual is probably present, discarding those where only sonar noise is present.

The average FP% on all the long clips is 0.67% for the DIDSON camera and 0.48% for the ARIS camera, meaning that an operator can expect a small rate of detections that are not fish. A more detailed analysis of the FP% reveals a large variance, especially for the DIDSON camera: the value, when calculated on the individual long clip, oscillates between 0.30 and 30, depending on the environmental conditions and/or the state of cleanliness of the lenses (on which the noise level depends). A more interesting indicator is therefore the median of the FP/TP rate. For the DIDSON data this metric is 0.83, while it is 0.40 for ARIS data (see Figure S.2 and S.3 for the distributions of the FT/TP rate). In both cases the median is below 1, suggesting that for most of the clips the number of FP will be smaller than the TP: with our model it is necessary a check of the detections by an operator, but in many cases it is a trivial operation.

4. Discussion

In a context of big data acquirement in ecological studies, such as camera trap data, the automation of the data extraction for analysis is one main issue that tends to be solved by the Convolutional Neural Networks (CNN) (Valletta et al., 2017). Indeed, those models are seeing a growing interest for their capacity to calculate the "good features" without any pre-treatment, but in case of complex data, such as sonar imaging, it yields to poor performances. A common route to solve this is to design or to modify an existing CNN architecture, for example adding more deep layers that are able to extract the peculiarity of the data as for noise treatment. However, this increases the number of parameters to optimize and yields to minimization difficulties. As a less common process, we suggest to pre-treat the data in order to elaborate an image composed by different masks, enhancing the signal. The advantages of this route are that it is easier to compare performances with existing ones developed in other fields, it can be based on the parameters and weights of other models even in complex situations and it is conceptually straightforward (opposed to over-complexed neural networks (Belkin et al., 2019)).

The protocol of validation must take into account the metrics used in deep learning to assess the performances of the algorithm. However real life application may deal with an amount of data with high levels of variability that might not be fully taken into account by standard deep learning validation protocols. Increasing the training dataset does not necessarily guarantee better results and requires considerable operator effort (Cho et al., 2016; C. Sun et al., 2017): the two-step validation that we proposed here can be considered as a viable strategy in order to limit the effort of labelling and validate the method on real-life scenarios. When the above conditions are met, we can therefore recommend trying the method outlined here at least as a first approach to exploring the limits of the most current CNNs.

The presented model itself exhibit interesting results. Looking at the TP% in function of fish size, we can affirm that the method is more than satisfactory (with values larger than ~75%) for all fish with size > 40 cm moving on DIDSON camera videos and > 60 cm for fish on ARIS. Despite the poor resolution of the acoustic camera images resolution face to optical images (Martignac et al., 2015), our method offers the possibility to optimize the detection of fish corresponding to these size thresholds without interfering with their behaviour with a transposable device on until 15-meters wide river sections, a clear advantage compared to the common monitoring devices such as video-counting.

Since we expect to apply such model to monitor fish populations in rivers where most of the recorded images will be of True Negative type (images where no a single fish is present) to reduce drastically the time an operator spend to search for a fish passage. With this in mind, our method proved to be remarkably efficient with both acoustic cameras, despite the differences in image quality and resolution. In parallel, we desired also a model able to minimize the FP rate, so that less time possible is spent in checking the correctness of the detections. Overall, the average FP% is low enough to reach our operational objectives, when compared to the total number of fish passages.

However some flaws are identified. The model show some difference in performances between DIDSON and ARIS cameras: for example, the ARIS TP% are always lower than DIDSON's (between ~5% and ~25% lower, see Table 3), for all fish sizes and species. This might be due a number of different aspects that differentiate the ARIS camera: a lower noise but higher image resolution, deeper record window resulting in smaller objects but also fewer images in the training set. A first possibility for improvement may be to increase the amount of ARIS data in the TD dataset, before moving on to develop two different models for the cameras. A second flaw is given by the different performances in function of the species of interest as, for example, for the European eel: the TP% is not optimal (below ~70% for 40-60 cm) neither with the DIDSON camera, nor with the ARIS (none of the values exceeds 70%). Indeed, the eels have a particular body shape and swimming behaviour, with a serpentine undulation (Webb, 1982) that make them less visible on the sonar images than the other fish species (Lagarde et al., 2020). Furthermore, European eel adults can be shorter than adults from another serpentine shape aquatic species like the Sea lamprey. Solving this requires to take into account explicitly the temporal correlation between images in which the same species is present. For example developing a LSTM-CNN based video object trackers as the ROLO architecture (Ning et al., 2017). Such kind of neural networks might help also to improve another characteristic: the low rate detections on small sized fishes, lower than 40 cm. The difficulty of some neural networks in detecting small objects is a well-known problem (Tong et al., 2020): an alternative method can be to test a CNN designed specifically for this kind of problem.

Beyond the performances, our model is characterized to have only one class "fish". This feature should not be considered a pitfall if the goal is a global monitoring of fish biological activity. Even if the datasets are elaborated from videos that were recorded on two sites, it has not to be excluded a certain degree of transferability of the model. In order to expand the method to more specifics uses (such as study of a single species), one of the most important steps in the future is, indeed, to improve this first version to increase the general efficiency and add multi-class identification.

To summarise, we developed a multi-camera and multi-species CNN-based neural network pipeline to automate fish recognition in a sonar video flux. We tried also to propose some recommendations in order to put to good use our experience such as the importance of pre-processing to keep models simple, the importance of a double validation or how to tackle cases of abundant TN data. Our model satisfy most of the required features, but this study only started exploring the potential of deep learning on such complex imaging and many challenges are still present. In a more general manner, the possibility to extract an accurate monitoring of fish populations from video-like analysis might answer to long-standing questions in aquatic ecology and open new ones, not only for researchers but also for biodiversity protection and management structures.

Acknowledgements

We gratefully thank the OFB (Office Français de la Biodiversité - French Office for Biodiversity) for the foundings.

Authors' contributions

G.F.G. developed the model, based on ideas developed with T.C.. M.M., F.M. and L.B. guided the development of the model concerning the ecological applications. F.M. collected the data and created the datasets along with G.F.G.. G.F.G. and F.M. wrote the first drafts and all authors discussed the results, implications and edited the manuscript.

Data Availability

The TD and VD datasets are available, respectively, at doi:10.5281/zenodo.5046708 and doi:10.5281/zenodo.5092010. The code is available at the following GitLab: https://forgemia.inra.fr/pole-migrateurs/Acoustic/acoustique.

References

Audebert, N., Saux, B. L., & Lefevre, S. (2019). Deep Learning for Classification of Hyperspectral Data : A Comparative Review. *IEEE Geoscience and Remote Sensing Magazine*, *7*(2), 159-173. https://doi.org/10.1109/MGRS.2019.2912563

Beery, S., Van Horn, G., & Perona, P. (2018). *Recognition in Terra Incognita*. 456-473. https://openaccess.thecvf.com/content_ECCV_2018/html/Beery_Recognition_in_Terra_ECCV_2018_paper.html

Belkin, M., Hsu, D., Ma, S., & Mandal, S. (2019). Reconciling modern machine-learning practice and the classical bias–variance trade-off. *Proceedings of the National Academy of Sciences*, *116*(32), 15849-15854. https://doi.org/10.1073/pnas.1903070116

Bonneau, M., Vayssade, J.-A., Troupe, W., & Arquet, R. (2020). Outdoor animal tracking combining neural network and time-lapse cameras. *Computers and Electronics in Agriculture*, *168*, 105150. https://doi.org/10.1016/j.compag.2019.105150

Chen, C. H. (2015). *Handbook Of Pattern Recognition And Computer Vision (5th Edition)*. World Scientific.

Cho, J., Lee, K., Shin, E., Choy, G., & Do, S. (2016). How much data is needed to train a medical image deep learning system to achieve necessary high accuracy? *arXiv:1511.06348 [cs]*. http://arxiv.org/abs/1511.06348

Christensen, J. H., Mogensen, L. V., & Ravn, O. (2020). Deep Learning based Segmentation of Fish in Noisy Forward Looking MBES Images. *IFAC-PapersOnLine*, *53*(2), 14546-14551. https://doi.org/10.1016/j.ifacol.2020.12.1459

Christin, S., Hervet, É., & Lecomte, N. (2019). Applications for deep learning in ecology. *Methods in Ecology and Evolution*, *10*(10), 1632-1644. https://doi.org/10.1111/2041-210X.13256

Cohen, J. (1960). A Coefficient of Agreement for Nominal Scales. *Educational and Psychological Measurement*, *20*(1), 37-46. https://doi.org/10.1177/001316446002000104

Cook, D., Middlemiss, K., Jaksons, P., Davison, W., & Jerrett, A. (2019). Validation of fish length estimations from a high frequency multi-beam sonar (ARIS) and its utilisation as a field-based measurement technique. *Fisheries Research*, *218*, 59-68. https://doi.org/10.1016/j.fishres.2019.05.004

Daum, D. W., & Osborne, B. M. (1998). Use of Fixed-Location, Split-Beam Sonar to Describe Temporal and Spatial Patterns of Adult Fall Chum Salmon Migration in the Chandalar River, Alaska. *North American Journal of Fisheries Management*, *18*(3), 477-486. https://doi.org/10.1577/1548-8675(1998)018<0477:UOFLSB>2.0.CO;2

Dougherty, E. R., & Lotufo, R. A. (2003). *Hands-on Morphological Image Processing*. SPIE Press.

Farley, S. S., Dawson, A., Goring, S. J., & Williams, J. W. (2018). Situating Ecology as a Big-Data Science : Current Advances, Challenges, and Solutions. *BioScience*, *68*(8), 563-576. https://doi.org/10.1093/biosci/biy068

Fleiss, J. L., Cohen, J., & Everitt, B. S. (1969). Large sample standard errors of kappa and weighted kappa. *Psychological Bulletin*, *72*(5), 323-327. https://doi.org/10.1037/h0028106

Foote, K. G. (2009). Acoustic Methods : Brief Review and Prospects for Advancing Fisheries Research. In R. J. Beamish & B. J. Rothschild (Éds.), *The Future of Fisheries Science in North America* (p. 313-343). Springer Netherlands. https://doi.org/10.1007/978-1-4020-9210-7_18

Girshick, R., Donahue, J., Darrell, T., & Malik, J. (2014). Rich feature hierarchies for accurate object detection and semantic segmentation. *ArXiv:1311.2524 [Cs]*. http://arxiv.org/abs/1311.2524

Guo, H. (2017). Big Earth data : A new frontier in Earth and information sciences. *Big Earth Data*, *1*(1-2), 4-20. https://doi.org/10.1080/20964471.2017.1403062

Helminen, J., & Linnansaari, T. (2021). Object and behavior differentiation for improved automated counts of migrating river fish using imaging sonar data. *Fisheries Research*, *237*, 105883. https://doi.org/10.1016/j.fishres.2021.105883

Huang, T., Yang, G., & Tang, G. (1979). A fast two-dimensional median filtering algorithm. *IEEE Transactions on Acoustics, Speech, and Signal Processing*, *27*(1), 13-18. https://doi.org/10.1109/TASSP.1979.1163188

Jianbo Shi & Tomasi. (1994). Good features to track. *1994 Proceedings of IEEE Conference on Computer Vision and Pattern Recognition*, 593-600. https://doi.org/10.1109/CVPR.1994.323794

Karim, F., Majumdar, S., Darabi, H., & Chen, S. (2018). LSTM Fully Convolutional Networks for Time Series Classification. *IEEE Access*, *6*, 1662-1669. https://doi.org/10.1109/ACCESS.2017.2779939

Keeken, O. A. van, Hal, R. van, Winter, H. V., Wilkes, T., & Griffioen, A. B. (2021). Migration of silver eel, Anguilla anguilla, through three water pumping stations in The Netherlands. *Fisheries Management and Ecology*, *28*(1), 76-90. https://doi.org/10.1111/fme.12457

Kocak, D. M., Dalgleish, F. R., Caimi, F. M., & Schechner, Y. Y. (2008). A Focus on Recent Developments and Trends in Underwater Imaging. *Marine Technology Society Journal*, *42*(1), 52-67. https://doi.org/10.4031/002533208786861209

Kwok, R. (2019). Deep learning powers a motion-tracking revolution. *Nature*, *574*(7776), 137-138. https://doi.org/10.1038/d41586-019-02942-5

Lagarde, R., Peyre, J., Amilhat, E., Mercader, M., Prellwitz, F., Simon, G., & Faliex, E. (2020). In situ evaluation of European eel counts and length estimates accuracy from an acoustic camera (ARIS). *Knowledge & Management of Aquatic Ecosystems*, *421*, 44. https://doi.org/10.1051/kmae/2020037

Landis, J. R., & Koch, G. G. (1977). The Measurement of Observer Agreement for Categorical Data. *Biometrics*, *33*(1), 159-174. https://doi.org/10.2307/2529310

Lankowicz, K. M., Bi, H., Liang, D., & Fan, C. (2020). Sonar imaging surveys fill data gaps in forage fish populations in shallow estuarine tributaries. *Fisheries Research*, *226*, 105520. https://doi.org/10.1016/j.fishres.2020.105520

LeCun, Y., Bengio, Y., & Hinton, G. (2015). Deep learning. *Nature*, *521*(7553), 436-444. https://doi.org/10.1038/nature14539

Lee, S., Park, B., & Kim, A. (2018). Deep Learning from Shallow Dives : Sonar Image Generation and Training for Underwater Object Detection. *ArXiv:1810.07990 [Cs]*. http://arxiv.org/abs/1810.07990

Lenihan, E. S., McCarthy, T. K., & Lawton, C. (2020). Assessment of silver eel (Anguilla anguilla) route selection at a water-regulating weir using an acoustic camera. *Marine and Freshwater Research*. https://doi.org/10.1071/MF20230

Long, J., Shelhamer, E., & Darrell, T. (2015). Fully convolutional networks for semantic segmentation. *2015 IEEE Conference on Computer Vision and Pattern Recognition (CVPR)*, 3431-3440. https://doi.org/10.1109/CVPR.2015.7298965

Lu, H., Li, Y., Zhang, Y., Chen, M., Serikawa, S., & Kim, H. (2017). Underwater Optical Image Processing : A Comprehensive Review. *Mobile Networks and Applications*, *22*(6), 1204-1211. https://doi.org/10.1007/s11036-017-0863-4

Ma, L., Liu, Y., Zhang, X., Ye, Y., Yin, G., & Johnson, B. A. (2019). Deep learning in remote sensing applications : A meta-analysis and review. *ISPRS Journal of Photogrammetry and Remote Sensing*, *152*, 166-177. https://doi.org/10.1016/j.isprsjprs.2019.04.015

Martignac, F., Daroux, A., Bagliniere, J.-L., Ombredane, D., & Guillard, J. (2015). The use of acoustic cameras in shallow waters : New hydroacoustic tools for monitoring migratory fish population. A review of DIDSON technology. *Fish and Fisheries*, *16*(3), 486-510. https://doi.org/10.1111/faf.12071

Maxwell, S. L., & Gove, N. (2002). *The Feasability of Estimating Migrating Salmon Passage Rates in Turbid Rivers using a Dual Frequency Identification Sonar (DIDSON)* (Regional Information Report Nº 2A04-05; p. 88). Alaska Department of Fish and Game.

Miele, V., Dussert, G., Spataro, B., Chamaillé-Jammes, S., Allainé, D., & Bonenfant, C. (2020). Revisiting giraffe photo-identification using deep learning and network analysis. *BioRxiv*, 2020.03.25.007377. https://doi.org/10.1101/2020.03.25.007377

Ning, G., Zhang, Z., Huang, C., Ren, X., Wang, H., Cai, C., & He, Z. (2017). Spatially supervised recurrent convolutional neural networks for visual object tracking. *2017 IEEE International Symposium on Circuits and Systems (ISCAS)*, 1-4. https://doi.org/10.1109/ISCAS.2017.8050867

Nixon, M., & Aguado, A. (2019). *Feature Extraction and Image Processing for Computer Vision*. Academic Press.

Norouzzadeh, M. S., Nguyen, A., Kosmala, M., Swanson, A., Palmer, M. S., Packer, C., & Clune, J. (2018). Automatically identifying, counting, and describing wild animals in camera-trap images with deep learning. *Proceedings of the National Academy of Sciences*, *115*(25), E5716-E5725. https://doi.org/10.1073/pnas.1719367115

Qi, C. R., Su, H., Mo, K., & Guibas, L. J. (2017). *PointNet : Deep Learning on Point Sets for 3D Classification and Segmentation*. 652-660. https://openaccess.thecvf.com/content_cvpr_2017/html/Qi_PointNet_Deep_Learning_CVPR_2017_paper.html

Raissi, M., Perdikaris, P., & Karniadakis, G. E. (2019). Physics-informed neural networks : A deep learning framework for solving forward and inverse problems involving nonlinear partial differential equations. *Journal of Computational Physics*, *378*, 686-707. https://doi.org/10.1016/j.jcp.2018.10.045

Redmon, J., Divvala, S., Girshick, R., & Farhadi, A. (2016). You Only Look Once : Unified, Real-Time Object Detection. *ArXiv:1506.02640 [Cs]*. http://arxiv.org/abs/1506.02640

Redmon, J., & Farhadi, A. (2018). YOLOv3 : An Incremental Improvement. *ArXiv*, 6.

Ren, S., He, K., Girshick, R., & Sun, J. (2016). Faster R-CNN : Towards Real-Time Object Detection with Region Proposal Networks. *ArXiv:1506.01497 [Cs]*. http://arxiv.org/abs/1506.01497

Salman, A., Siddiqui, S. A., Shafait, F., Mian, A., Shortis, M. R., Khurshid, K., Ulges, A., & Schwanecke, U. (2020). Automatic fish detection in underwater videos by a deep neural network-based hybrid motion learning system. *ICES Journal of Marine Science*, *77*(4), 1295-1307. https://doi.org/10.1093/icesjms/fsz025

Schneider, S., Taylor, G. W., & Kremer, S. (2018). Deep Learning Object Detection Methods for Ecological Camera Trap Data. *2018 15th Conference on Computer and Robot Vision (CRV)*, 321-328. https://doi.org/10.1109/CRV.2018.00052

Shahrestani, S., Bi, H., Lyubchich, V., & Boswell, K. M. (2017). Detecting a nearshore fish parade using the adaptive resolution imaging sonar (ARIS) : An automated procedure for data analysis. *Fisheries Research*, *191*, 190-199. https://doi.org/10.1016/j.fishres.2017.03.013

Sun, C., Shrivastava, A., Singh, S., & Gupta, A. (2017). *Revisiting Unreasonable Effectiveness of Data in Deep Learning Era*. 843-852.

https://openaccess.thecvf.com/content_iccv_2017/html/Sun_Revisiting_Unreasonable_Effectiveness_ICCV_2017_paper.html

Sun, R. (2019). Optimization for deep learning : Theory and algorithms. *ArXiv:1912.08957 [Cs, Math, Stat]*. http://arxiv.org/abs/1912.08957

Tarling, P., Cantor, M., Clapés, A., & Escalera, S. (2021). Deep learning with self-supervision and uncertainty regularization to count fish in underwater images. *ArXiv:2104.14964 [Cs]*. http://arxiv.org/abs/2104.14964

Tong, K., Wu, Y., & Zhou, F. (2020). Recent advances in small object detection based on deep learning : A review. *Image and Vision Computing*, *97*, 103910. https://doi.org/10.1016/j.imavis.2020.103910

Valdenegro-Toro, M. (2016). *Submerged Marine Debris Detection with Autonomous Underwater Vehicles*. 7.

Valletta, J. J., Torney, C., Kings, M., Thornton, A., & Madden, J. (2017). Applications of machine learning in animal behaviour studies. *Animal Behaviour*, *124*, 203-220. https://doi.org/10.1016/j.anbehav.2016.12.005

Wang, H., & Raj, B. (2017). On the Origin of Deep Learning. *ArXiv:1702.07800 [Cs, Stat]*. http://arxiv.org/abs/1702.07800

Wang, Y., Song, W., Fortino, G., Qi, L., Zhang, W., & Liotta, A. (2019). An Experimental-Based Review of Image Enhancement and Image Restoration Methods for Underwater Imaging. *IEEE Access*, *7*, 140233-140251. https://doi.org/10.1109/ACCESS.2019.2932130

Webb, P. (1982). Locomotor Patterns in the Evolution of Actinopterygian Fishes. *American Zoologist*, *22*(2), 329-342. https://doi.org/10.1093/icb/22.2.329

Weinstein, B. G. (2018). A computer vision for animal ecology. *Journal of Animal Ecology*, *87*(3), 533-545. https://doi.org/10.1111/1365-2656.12780

Xia, K., Huang, J., & Wang, H. (2020). LSTM-CNN Architecture for Human Activity Recognition. *IEEE Access*, *8*, 56855-56866. https://doi.org/10.1109/ACCESS.2020.2982225

Yang, Y.-S., Bae, J.-H., Lee, K.-H., Park, J.-S., & Sohn, B.-K. (2010). Fish Monitoring through a Fish Run on the Nakdong River using an Acoustic Camera System. *Korean Journal of Fisheries and Aquatic Sciences*, *43*(6), 735-739. https://doi.org/10.5657/kfas.2010.43.6.735

Zacchini, L., Franchi, M., Manzari, V., Pagliai, M., Secciani, N., Topini, A., Stifani, M., & Ridolfi, A. (2020). Forward-Looking Sonar CNN-based Automatic Target Recognition : An experimental campaign with FeelHippo AUV. *2020 IEEE/OES Autonomous Underwater Vehicles Symposium (AUV)*, 1-6. https://doi.org/10.1109/AUV50043.2020.9267902

Zdeborová, L. (2020). Understanding deep learning is also a job for physicists. *Nature Physics*, *16*(6), 602-604. https://doi.org/10.1038/s41567-020-0929-2

Zivkovic, Z., & van der Heijden, F. (2006). Efficient adaptive density estimation per image pixel for the task of background subtraction. *Pattern Recognition Letters*, *27*(7), 773-780. https://doi.org/10.1016/j.patrec.2005.11.005

# Supporting Information

|  | DIDSON Std. 300 m |  | ARIS Explorer 3000 |  |
| --- | --- | --- | --- | --- |
| Mode | Low Frequency | High Frequency | LF | HF |
| Frequency (MHz) | 1.1 | 1.8 | 1.8 | 3.0 |
| Number of beams | 48 | 96 | 128 | 128 |
| Range bin number | 512 | 512 | 512 | 512 |
| Beamwidth (two-way) | 0.4° H by 14° V | 0.3° H by 14° V | 0.3° H by 15° V | 0.2° H by 15° V |
| Beam spacing | 0.60° | 0.30° | 0.25° | 0.25° |
| Overall field of view | 29° H by 14° V | 29° H by 14° V | 30° H by 15° V | 30° H by 15° V |
| Max frame rate | 4-21 frames/sec. | 4-21 frames/sec. | 4-15 frames/sec. | 4-15 frames/sec. |
| Window Length | 5 to 40 m | 1.25 to 10 m | Up to 15 m | Up to 5 m |
| Transmit pulse length | 18 µs to 144 µs | 4.5 µs to 36 µs | 4 µs to 24 µs | 4 µs to 24 µs |

Table S.1: Comparison of key characteristics of acoustic cameras DIDSON 300 and ARIS 3000.

S.1 Datasets Details

All the species, which are present both datasets, may be target species on ecological monitoring studies in aquatic environments. Most monitoring of diadromous fish population with acoustic cameras focus on large migratory salmonids, such as Atlantic salmons, or European eels, that are two of the most endangered fish in North-West of France. Consequently, those two species are over-represented.

In the TD dataset, all the images are extracted from 23 ARIS clips and 39 DIDSON clips. The TD is divided into a training set (60% of the images), an evaluation set (19%) and a test set (21%): in this scheme, typical in deep learning, "train" data is used to optimise the CNN parameters, "val" data to check the metrics during the optimisation and "test" data to calculate the metrics once the optimisation is over.

In the VD dataset, the total dataset correspond to around 660 000 images.

| Species | Total clips | Total annotations | ARIS clips | ARIS annotations | DIDSON clips | DIDSON annotations |
|---|---|---|---|---|---|---|
| Atlantic salmon | 20 | 1117 | 6 | 268 | 14 | 849 |
| European eel | 12 | 1161 | 8 | 568 | 4 | 593 |
| Sea lamprey | 9 | 324 | 1 | 22 | 8 | 324 |
| Allis shad | 11 | 718 | 1 | 290 | 10 | 428 |
| Giant catfish | 10 | 633 | 7 | 418 | 3 | 152 |
| Total | 62 | 3953 | 23 | 1566 | 39 | 2346 |

Table S.2 : distribution of each species of interest in the 50 clips of the TD. Each clip may contain more than one species and more than one passage.

| Species | Camera | Fish size | | | | Total |
| | | 20-40 | 40-60 | 60-80 | > 80 | |
|---|---|---|---|---|---|---|
| Fish species of main interest | | | | | | |
| Atlantic salmon | DIDSON | 0 | 14 | 33 | 6 | 53 |
| | ARIS | 0 | 2 | 10 | 0 | 12 |
| European eel | DIDSON | 0 | 14 | 15 | 3 | 32 |
| | ARIS | 13 | 52 | 15 | 2 | 82 |
| Other fish species, included | | | | | | |
| Generic Fish (unidentified) | DIDSON | 315 | 505 | 30 | 2 | 852 |
| | ARIS | 248 | 419 | 18 | 0 | 685 |
| Sea lamprey | DIDSON | 0 | 0 | 2 | 2 | 4 |
| | ARIS | 0 | 0 | 0 | 1 | 1 |
| Allis shad | DIDSON | 0 | 1 | 0 | 0 | 1 |
| | ARIS | 0 | 0 | 0 | 0 | 0 |
| European catfish | DIDSON | 0 | 0 | 0 | 2 | 2 |
| | ARIS | 0 | 0 | 0 | 6 | 6 |

Table S.3: Number of clips for the species of interest by size in the VD dataset.

| Metric | Equation | Notes |
|---|---|---|
| Precision (P) | $P = \dfrac{TP}{TP+FP}$ | TP is the total number of True Positive and FP of False Positive. P is a measure of how many relevant data are found, compared to the whole found data. |

| Recall (R) | $R = \dfrac{TP}{TP+FN}$ | FN is the total number of False Negative. R is a measure of how many relevant data are found, compared to the whole relevant data. |
|---|---|---|
| $F_1$-score | $F_1 = 2\dfrac{P \times R}{P+R}$ | $F_1$ is a mean between P and R. Its are between 0 and 1 (perfect recall and precision). |
| Intersection over Union (IoU) | $IoU = \dfrac{area(bb_p \cap bb_{g-t})}{area(bb_p \cup bb_{g-t})}$ | $bb_p$ is a predicted bounding box and $bb_{g-t}$ is a ground-truth box. The IoU, used in the calculation of the AP@0.50, is a measure of the spatial agreement between the ground-truth and predicted boxes. |
| Average Precision (AP@0.50) | $AP = \sum_{i}^{n-1}(r_{i+1}-r_i)p(r_{i+1})$ | $r_i$ are the recall levels and $p(r_i)$ is the interpolated precision. See PASCAL development kit for more details. Only the detections with an IoU ≥ 0.50 are counted in the calculation of the AP. This metric, commonly used in deep learning, is the average precision between different recall levels. |
| Cohen's kappa (κ) | $\kappa = \dfrac{p_0 - p_e}{1 - p_e}$ | κ is a measure of the inter-rater reliability. In our case the first rater is the neural network and the second is the operator that annotated the ground-truth. $p_0$ is the observed agreement between raters. $P_e$ is the probability of a random agreement. |

Table S.4: Metrics used for the model validation.

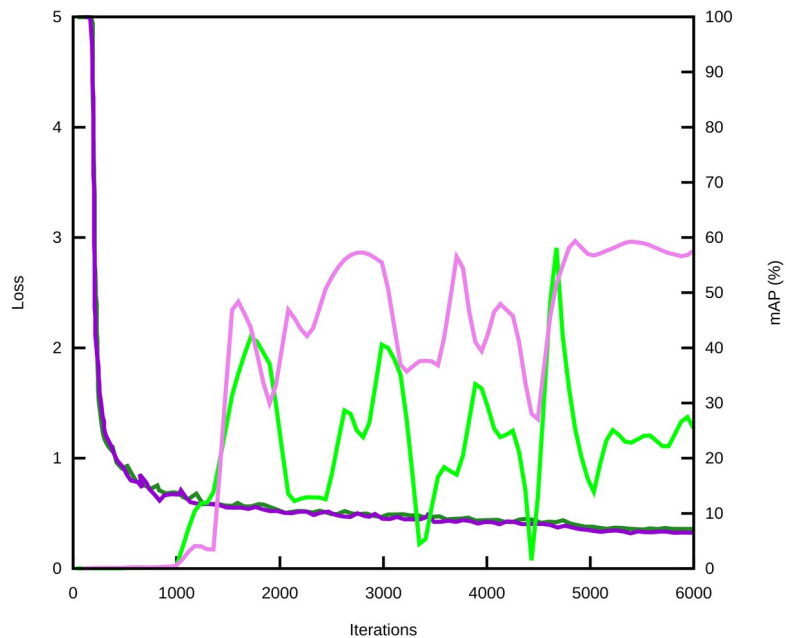

Figure S.1: training run with only the 'b' mask (violet) and only the '$b_f$' mask (green). In dark colors the loss function and in light colors the AP@0.50 as a validation metric.

| DIDSON | Fish | no Fish |
|---|---|---|
| Fish | 0.73 | 0.05 |
| no Fish | 0.27 | 0.95 |

| ARIS | Fish | no Fish |
|---|---|---|
| Fish | 0.45 | 0.24 |
| no Fish | 0.55 | 0.76 |

Table S.5 : Confusion matrices calculated on the « test » set of the TD dataset for DIDSON (left) and ARIS (right) cameras.

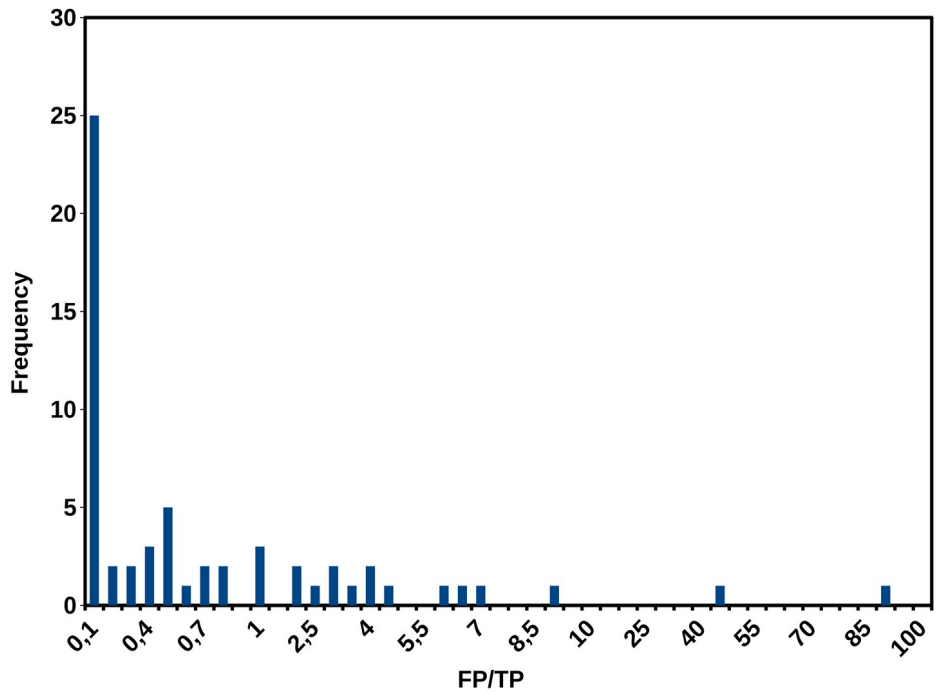

Figure S.2 : FT/TP histogram for ARIS camera on the VD dataset ; to highlight the data, the scale has increments of 0.1 between 0.0 and 1.0, 1.5 between 1.0 and 10, 5 between 10 and 100.

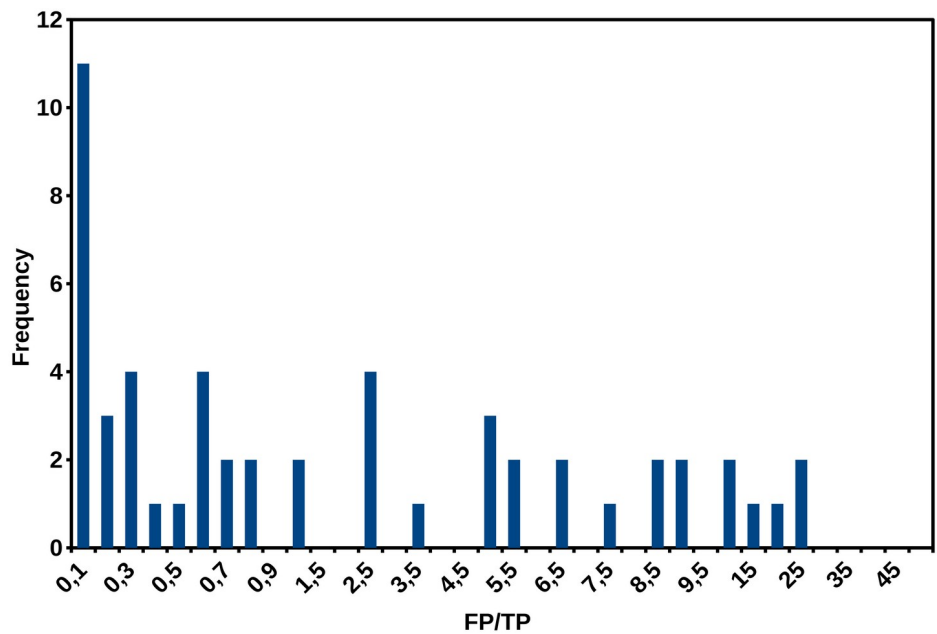

Figure S.3 : FT/TP histogram for DIDSON camera on the VD dataset ; to highlight the data, the scale has increments of 0.1 between 0.0 and 1.0, 1.5 between 1.0 and 10, 5 between 10 and 50.